\documentclass[preprint,12pt]{elsarticle}

\usepackage[utf8]{inputenc}
\usepackage[T1]{fontenc}
\usepackage{lmodern}
\usepackage{graphicx}
\usepackage{amsmath,amsfonts,amssymb}
\usepackage{booktabs}
\usepackage{url}
\usepackage[hidelinks]{hyperref}

\begin{document}

\begin{frontmatter}

\title{Data collection from highways: a geometric, class-agnostic approach\\ to embedded vehicle counting\tnoteref{t1}}

\tnotetext[t1]{Retrospective preprint. The prototype, the experiments and the field
deployment reported in Sections~\ref{sec:pipeline}--\ref{sec:results} were carried out
between 2016 and 2018 as a final-year engineering project at the Universidade Federal de
Alagoas, and are reported here essentially unchanged. Sections~\ref{sec:related},
\ref{sec:observations} and \ref{sec:retrospect} were written later and place the method in
the context of current practice, in which background subtraction survives as a
\emph{class-agnostic} detector for the many settings where no trained model exists for the
object class of interest.}

\author[mycorrespondingauthor]{Lucas Gouveia Omena Lopes}
\cortext[mycorrespondingauthor]{Corresponding author}
\author{William W. M. Lira}
\author{Alexandre M. Lima}
\author{Thales M. A. Vieira}
\address{Universidade Federal de Alagoas, Maceió, Alagoas, Brazil}

\begin{abstract}
Traffic data collection is dominated today by deep object detectors followed by
tracking-by-detection, a pipeline that presupposes what is often missing in practice: a
detector already trained on the class one wants to count. We revisit a purely geometric
traffic-sensing pipeline for Single Board Computers in which detection is
\emph{class-agnostic}: moving objects come from background subtraction and thresholding,
and counting is decided by a geometric rule on an imaginary line across the road, a
software inductive loop detector. With no object model, training set or per-object
trajectory, it runs faster than real time on Raspberry Pi class hardware. Two counting
rules are described: a \emph{constant average speed} rule, whose expected accuracy is
derived analytically as $\approx\!86\%$ under a Gaussian speed distribution, and a
self-calibrating \emph{pre-calibration} rule that recovers the lane geometry from blob
statistics and counts edges of lane occupancy, additionally yielding per-vehicle average
speed at no extra cost. Over four videos the latter counts with $83.3\%$--$100\%$
accuracy; in a field deployment it reaches $91\%$ against $37.5\%$ for a blob-tracking
baseline under the same compute budget. We report the observations of that period in
detail: the resolution floor below which accuracy collapses, the frame rate floor at which
vehicles alias past the counting line, the gap between short curated clips and long
uncontrolled footage, and the trade-off between Python (easier to tune, $100\%$ CPU) and
C++ ($40\%$ CPU, thermally viable). These are properties of the sampling geometry, not of
the hardware of the time, and still constrain edge deployments. We close by arguing where
motion-based, class-agnostic detection remains the right tool: open-set classes with no
annotated data, tight power budgets, privacy-constrained installations, and the cold start
of mining training crops to bootstrap a learned detector.
\end{abstract}


\end{frontmatter}

\section{Introduction}
\label{sec:intro}

Decisions in traffic engineering are supported by simulation models whose main input is
the origin--destination demand matrix, which states how many people or vehicles wish to
travel between each pair of origin and destination points of a
network~\cite{ortuzar2011}. That matrix is estimated, in large part, from vehicle counts
on the links of interest~\cite{vanzuylen1980,cascetta1984}, which makes volumetric
counting a permanent, recurring need of urban transport planning rather than a one-off
measurement. The cost of obtaining those counts --- whether by human observers, by
pneumatic tubes or by inductive loops buried in the pavement --- is what limits how
densely and how often a network can be instrumented.

Embedded systems offer an attractive alternative because they consume little energy and
little space, are designed for a single function and therefore demand no specialised
knowledge from the operator. In this work the prototype is built on Single Board Computers
(SBCs): complete computers on one board, cheap and small, able to run a general-purpose
operating system and consequently to capture and process video, unlike conventional
microcontrollers. Image capture and processing on such a device turn a fixed camera into a
traffic sensor.

\subsection{The problem this paper actually addresses}

At the time this system was designed, the works dealing with the extraction of traffic
information from images fell into two groups. One group pursued accuracy at the cost of
computation: Gupte et al.~\cite{gupte2002} detect, track and classify vehicles from a fixed
camera, recovering position, dimensions and speed; Song and Nevatia~\cite{song2005}
introduce a vehicle model to constrain segmentation and tracking, so as to avoid errors in
the extracted data; Hinz~\cite{hinz2003} treats a further set of radiometric and geometric
factors that raise detection accuracy in aerial imagery. These solutions require
conventional computers. The other group traded accuracy for portability, building counting
systems out of a short chain of classical operators so that they could run on commodity
machines~\cite{tannus2012} or on user-defined virtual loops~\cite{barcellos2015} --- but
without reporting an embedded platform on which the method actually lives.

What both groups share is the assumption that detection must first produce
\emph{object-level} evidence --- a segmented vehicle, a trajectory, a class label --- and
that counting is a consequence of that evidence. This paper takes the opposite route. It
proposes a geometric system, running in real time on an embedded device, in which the
quantity of interest is measured directly on the image geometry, and object identity is
never established. Detection is \emph{class-agnostic}: anything that moves against the
learned background and satisfies a few shape constraints is a candidate; counting is
decided by whether such a candidate occupies a thin band drawn across the road. This is
the software equivalent of the inductive loop~\cite{michalopoulos1991}, and it is why the
pipeline fits in an SBC.

\subsection{Why revisit this now}
\label{sec:why-now}

The obvious objection is that the problem is solved: a modern detector such as
YOLO~\cite{redmon2016yolo} or Faster R-CNN~\cite{ren2015fasterrcnn}, followed by
tracking-by-detection~\cite{bewley2016sort,wojke2017deepsort,zhang2022bytetrack}, counts
vehicles more accurately than any geometric heuristic, and runs on today's accelerated
edge hardware. That objection holds \emph{whenever a trained detector for the target class
exists}. The premise is stronger than it looks. Learned detectors are trained on closed
vocabularies --- COCO~\cite{lin2014coco} contains \texttt{car}, \texttt{bus},
\texttt{truck}, \texttt{motorcycle}, and nothing else that moves on a road. The moment the
object of interest falls outside that vocabulary --- animal crossings, debris and lost
load, agricultural and construction machinery, informal transport modes, carts, unusual
oversized loads, or simply ``any object that entered the restricted lane'' --- there is no
model and there is no annotated data with which to train one. Open-vocabulary detectors
\cite{radford2021clip,liu2024groundingdino} and promptable segmenters
\cite{kirillov2023sam,ravi2024sam2} relax the vocabulary constraint, but they relax it at a
computational cost that is orders of magnitude above the budget of a passively cooled board
on a pole, and their behaviour on narrow, domain-specific classes is not guaranteed.

Background subtraction is not a competitor to learned detection; it is what one uses when
learned detection is not available. Its defining property is exactly the one the deep
pipeline lacks: it needs no class model at all, because it defines the foreground by what
changes rather than by what is recognised. In our own practice this is still how tracking
is bootstrapped whenever there is no trained model for the classes at hand --- as a
class-agnostic detector feeding a standard association stage, and as an automatic proposal
generator that mines the crops later used to train the model that will eventually replace
it. Section~\ref{sec:retrospect} develops this argument; the rest of the paper is the
original system and, above all, the empirical observations it produced.

\subsection{Contributions}

\begin{enumerate}
\item A tracking-free, class-agnostic counting formulation in which the geometric decision
is moved from the object to a fixed line on the road, reducing per-frame cost from
$O(\text{objects} \times \text{history})$ to $O(1)$ tests
(Section~\ref{sec:method}).
\item An analytical error model for the constant-average-speed rule, giving an expected
counting accuracy of $\approx\!86\%$ under a Gaussian speed distribution, and the
resulting sampling condition that relates band thickness, frame rate and pixel--metre
scale (Section~\ref{sec:alg1}).
\item A self-calibrating variant that infers lane boundaries from the statistics of the
detected blobs and counts occupancy edges per lane, with a by-product estimate of
per-vehicle average speed (Sections~\ref{sec:alg2} and \ref{sec:speed}).
\item A characterisation across four SBC platforms, four videos and one field deployment,
and an explicit account of the failure modes observed
(Sections~\ref{sec:results} and \ref{sec:observations}).
\item A retrospective positioning of motion-based, class-agnostic detection with respect
to current detector-based practice, including the hybrid recipe we still use
(Section~\ref{sec:retrospect}).
\end{enumerate}

The prototype is validated on volumetric vehicle counting, which is one instance of a
broader family of problems --- axle counting, per-lane average speed, queue length at
signals --- that the same geometric machinery addresses with minor adaptations.

\section{Related work and historical position}
\label{sec:related}

\subsection{The virtual loop lineage}

Counting by testing a fixed region of the image, rather than by following objects, predates
this work by decades. The Autoscope system~\cite{michalopoulos1991} replaced buried
inductive loops with user-drawn detection zones in the camera image, and
Beymer et al.~\cite{beymer1997} and Coifman et al.~\cite{coifman1998} built real-time
traffic-parameter systems around fixed cameras and feature-level evidence. The strategy in
Section~\ref{sec:method} belongs to this lineage: the imaginary line is a virtual loop, and
its main design freedom is its thickness, which we tie explicitly to the sampling rate.

\subsection{Background modelling}

The foreground extraction used here is the simplest member of a long family. Running-average
background models were superseded by per-pixel mixtures of Gaussians~\cite{stauffer1999},
refined for faster convergence and shadow handling~\cite{kaewtrakulpong2001} and given an
adaptive number of components~\cite{zivkovic2004}; sample-consensus models such as
ViBe~\cite{barnich2011vibe} and feedback-driven methods such as
SuBSENSE~\cite{stcharles2015subsense} improved robustness to illumination change and
dynamic backgrounds; the field is surveyed in~\cite{bouwmans2014} and benchmarked on
CDnet~\cite{wang2014cdnet}. All of these are drop-in replacements for the subtraction stage
of Section~\ref{sec:bgs} and none of them changes the counting logic, which is the point:
the geometric decision layer is agnostic to how the foreground mask was produced.

\subsection{What changed afterwards}

Since the experiments reported here, detection moved decisively to learned models
\cite{ren2015fasterrcnn,redmon2016yolo} and multi-object tracking converged on
tracking-by-detection, where a detector produces boxes per frame and an association stage
links them over time --- with motion only~\cite{bewley2016sort,bochinski2017iou}, with
appearance embeddings~\cite{wojke2017deepsort}, or by associating low-confidence boxes as
well~\cite{zhang2022bytetrack}. Evaluation was standardised by the MOT
benchmarks~\cite{milan2016mot16} and metrics~\cite{bernardin2008clearmot,luiten2021hota},
and traffic-specific evaluation by the AI City Challenge~\cite{naphade2021aicity}.

The relevant observation for this paper is structural rather than historical: the
association stage of every one of those trackers is indifferent to the origin of the
detections. Replacing the detector with a background-subtraction blob extractor yields a
class-agnostic tracker that needs no training data, which is precisely the configuration we
still resort to when no model exists for the class in question.

\section{The class-agnostic detection pipeline} \label{sec:pipeline}

This section describes the front end that turns a video stream into foreground blobs. Every
stage is a classical, analytical operation with a fixed cost per pixel; none of them
involves learning, and none of them assumes anything about \emph{what} the moving object
is. This is what makes the front end reusable outside the traffic domain, and it is the
reason the same pipeline is still serviceable when no trained model is available for the
class of interest.

Computer vision is the analysis of images to collect information in the way human vision
does~\cite{szeliski2022}, and its applications --- face recognition for access control
in industrial and construction sites, optical character recognition for document
digitalisation~\cite{mori1992} and licence-plate reading~\cite{du2013alpr},
motion-triggered surveillance recording --- each rely on a specific selection of
image-processing operators. The selection is dictated by cost. An image of only
$320\times240$ pixels requires at least $76\,800$ iterations for a single simple filter,
and that figure multiplies by frame rate and by the number of filters in the chain.
Feasibility, not only correctness, drives the design. The following subsections present the
operators used, each chosen for a low and predictable cost.

\subsection{Image resizing}
\begin{figure*}[h!]
\centering
\includegraphics*[width=0.5\textwidth]{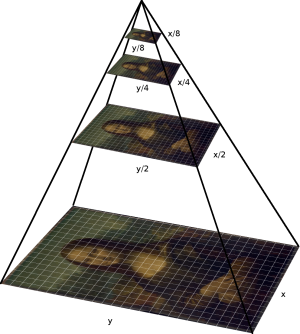}
\caption{An example of an image pyramid~\cite{Rosebrock2015}.}
\label{fig:red}
\end{figure*}

Resolution describes the level of detail an image contains; the HD label 720p, for instance,
denotes $1280\times720$ pixels. Since processing cost is directly proportional to the number
of pixels, the captured frames are resized before anything else in the chain. Resizing maps
the image onto a different resolution while preserving proportions, with new pixels obtained
by interpolating the originals; the implementations used here come from the OpenCV
library~\cite{bradski2000opencv}. Fig.~\ref{fig:red} illustrates an image pyramid, a
multi-scale representation of the same image at successive resolutions.

Resizing is the single most effective knob in the whole system, and also the most dangerous:
Section~\ref{sec:obs-resolution} shows that counting accuracy is flat above a resolution
threshold and collapses below it, so the operating point should be chosen just above that
knee rather than as low as the hardware tolerates.

\subsection{Grayscale conversion}
An image is a scalar field $f(x,y)$ over Cartesian coordinates, each coordinate pair holding
a picture element, or pixel. Colour is encoded by patterns such as HSI, CMYK or, most
commonly, RGB, in which each pixel carries three intensities and from whose combination most
colours visible to the human eye can be composed~\cite{gonzalez2008}. When the quantity of
interest is shape and motion rather than colour, the image is converted to grayscale, where
each pixel carries only monochromatic luminous intensity, conventionally in the range
$[0,255]$. The conversion from the intensities $(R_{xy},G_{xy},B_{xy})$ of pixel $(x,y)$ into
a gray intensity $C_{xy}$ follows the luminance expression of ITU-R
Recommendation~BT.601~\cite{itu2011bt601}:

\begin{equation}
    C_{xy} = 0.299\,R_{xy} + 0.587\,G_{xy} + 0.114\,B_{xy}.
\end{equation}

Besides the threefold reduction in memory traffic, discarding colour removes a nuisance
variable: on an outdoor road scene the chromatic content varies with the time of day far
more than the luminance contrast between vehicle and pavement does.

\subsection{Gaussian blur}
Gaussian functions are exponentials whose graph has the shape of a bell, the normal
distribution curve. Gaussian smoothing convolves the image with such a
kernel~\cite{gonzalez2008}, reducing definition and, more importantly, noise. Noise removal
matters here because noise appears as abrupt local changes, which is exactly the signature
the subtraction stage is looking for; without smoothing, sensor noise is indistinguishable
from a small moving object.

\subsection{Background subtraction and thresholding}
\label{sec:bgs}
In OpenCV a grayscale image is a two-dimensional matrix (rather than the three-dimensional
structure of the RGB frame it came from) whose entry $(i,j)$ is the intensity of the
corresponding pixel. This representation admits arithmetic operations such as subtraction
and addition, whose result is another matrix, that is, another image. A video is a
succession of such images, called frames, and from the information contained across frames
one can estimate the \emph{background}: the part of the scene that does not move. Subtracting
the background from any frame leaves an image that contains only the moving objects.

In principle the background is a frame in which no other object is present. In practice such
a frame rarely exists, and both the illumination and the static contents of the scene change
over time. The estimate used here therefore initialises the background with the first frame
and updates it by weighted interpolation with the subsequent ones, a running-average model.
This is the cheapest member of the family surveyed in Section~\ref{sec:related}: mixtures of
Gaussians~\cite{stauffer1999,zivkovic2004}, sample consensus~\cite{barnich2011vibe} or
feedback-driven models~\cite{stcharles2015subsense} may be substituted here without touching
anything downstream, trading cycles for robustness.

\begin{figure*}[h!]
\centering
\includegraphics*[width=0.9\textwidth]{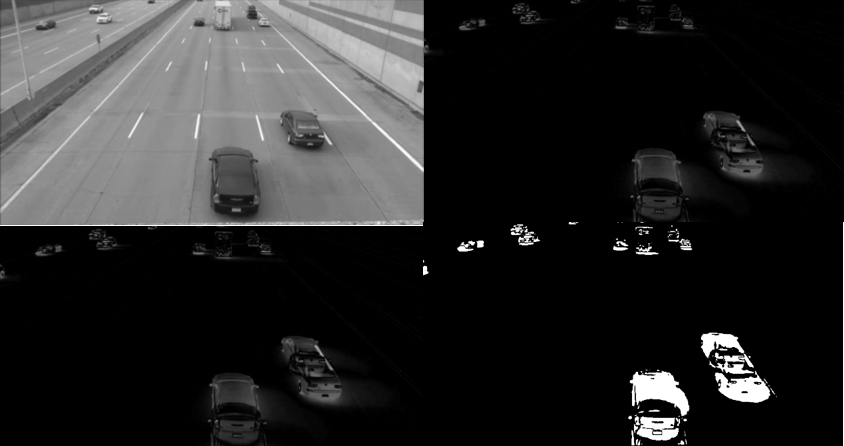}
\caption{Subsequent application of background extraction and thresholding.}
\label{fig:bt}
\end{figure*}

The difference image is never clean. Illumination shifts between frames, focus changes and
other factors make the true background drift subtly from one frame to the next, so the
subtraction leaves, besides the moving objects, a residue of faintly lit background.
Thresholding removes it: every pixel whose intensity is above a threshold is set to the
maximum value (white, $255$) and every other to the minimum (black, $0$), yielding a binary
matrix~\cite{gonzalez2008}. The residue is erased and the moving objects are isolated.
Fig.~\ref{fig:bt} shows both steps applied in sequence.

\subsection{Dilation and contour extraction}
Dilation thickens the objects of a binary image, with the shape and extent of the growth
controlled by the structuring element. Its purpose here is to reconnect parts of the same
object: after thresholding, pixels belonging to a moving object may have fallen below the
threshold, leaving the object fragmented or perforated. Dilation closes those gaps so that a
vehicle is recovered as one blob rather than several.

\begin{figure*}[h!]
\centering
\includegraphics*[width=0.9\textwidth]{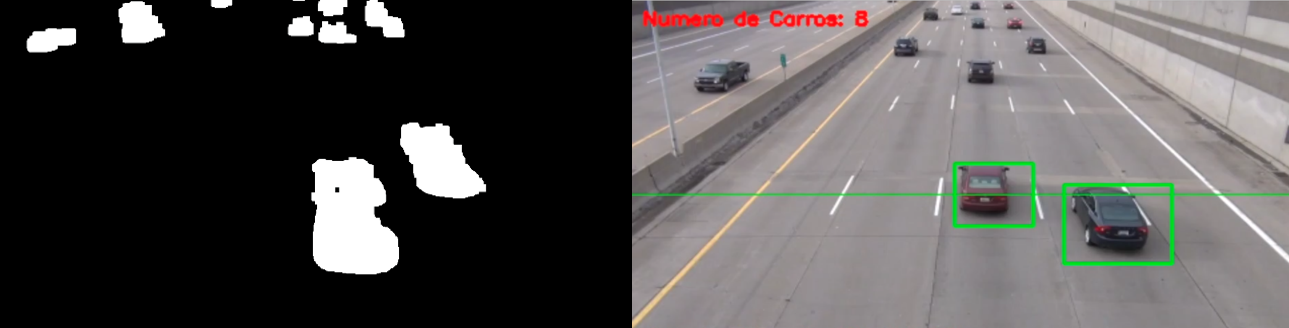}
\caption{Subsequent application of dilation and contour extraction.}
\label{fig:dc}
\end{figure*}

From the dilated mask, the contours of the pixel clusters are extracted with the OpenCV
contour-tracing procedure, and filtered by the parameters of interest --- minimum area,
maximum area, and length-to-width ratio. Each surviving contour is approximated by its
bounding rectangle, which yields a centre, a width and a height. These few scalars are the
entire object description the system ever uses: no appearance model, no class label, no
identity. Fig.~\ref{fig:dc} shows the two steps applied to the result of Fig.~\ref{fig:bt}.

The shape filters are worth an explicit remark, because they are the only place where prior
knowledge about the object enters the system. They encode a weak, geometric prior --- ``an
object of roughly this apparent size and elongation'' --- not a class model. Retargeting the
pipeline to a different kind of object is a matter of changing two or three numbers, which
is exactly the property that a trained detector does not have.

\section{Cost analysis of tracking-based counting}
\label{sec:baseline}

This section quantifies why object-level tracking was not viable on the target hardware,
which is the observation that motivated the geometric formulation of
Section~\ref{sec:method}.

\subsection{The tracking-based baseline}
The detection strategies of the period --- from the classical pipelines of
Gupte et al.~\cite{gupte2002} and Tannús et al.~\cite{tannus2012} to the widely reproduced
open-source implementation of Dahms~\cite{dahms2016} --- share the pre-processing and
background-subtraction stages described above, and differ from each other mainly in how
counting is decided. They store information from consecutive frames in order to predict the
next stage of the motion of each contour; knowing the predicted position of each contour,
they test it against an imaginary line, and a contour that crosses the line is added to the
volumetric count. This is the \emph{basic blob tracking} strategy, and its cost is dominated
by maintaining and predicting the state of every contour in the scene.

\begin{figure*}[h!]
\centering
\includegraphics*[width=0.9\textwidth]{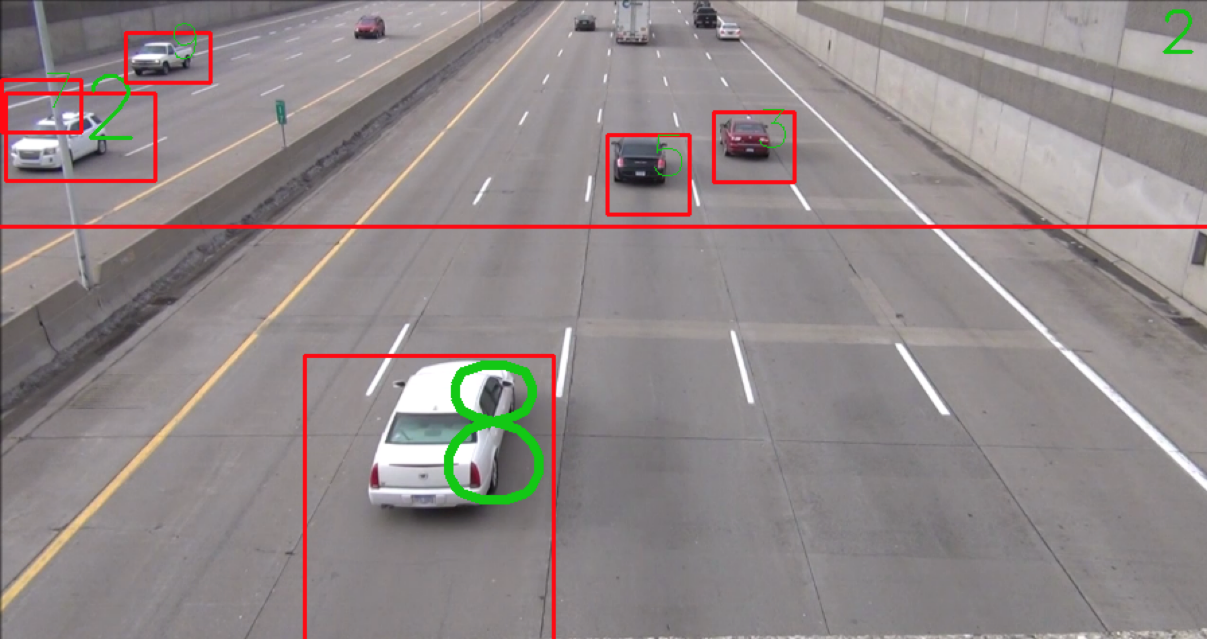}
\caption{Volumetric vehicle counting using the blob-tracking baseline~\cite{dahms2016}.}
\label{fig:dahms}
\end{figure*}

The approach is conceptually close to intrusive analog detection, with a line acting as the
imaginary obstacle, and the tracking stage exists for one reason: to guarantee that the same
vehicle is not counted twice. It is a sound strategy and it is expensive --- expensive enough
that contemporary attempts to move it to portable hardware had to shrink the input to
resolutions of the order of $160\times120$ and restrict the blob search to the band where
vehicles actually move, and still processed video far slower than real time.

We quantified the effect directly with the open-source C++ implementation released by
Dahms~\cite{dahms2016}, on the 32\,s, 720p, 30\,fps clip distributed with it. Resolution and
frame rate were reduced across a range of values and the response measured.
Fig.~\ref{fig:dahms} shows a processed frame. Results are in Table~\ref{tab:dahms}; the
ground-truth count for this clip is $51$ vehicles.

\begin{table}[ht]
  \begin{center}
    \caption{Blob-tracking baseline~\cite{dahms2016} on a 32\,s clip, on a 2.4\,GHz Intel
    Core i5 notebook with 8\,GB RAM. Ground-truth count: 51 vehicles. Note that no
    configuration processes the clip faster than $32$\,s except the most degraded ones,
    and those are also the least accurate.}
    \label{tab:dahms}
    \begin{tabular}{cccc}
      \toprule
      \textbf{Resolution} & \textbf{FPS} & \textbf{Processing time} & \textbf{Count}\\
      \midrule
        720p & 30 & 177\,s & 51\\
        480p & 30 & 105\,s & 59\\
        120p & 30 & 42\,s & 30\\
        720p & 15 & 73\,s & 55\\
        480p & 15 & 46\,s & 49\\
        120p & 15 & 27\,s & 14\\
      \bottomrule
    \end{tabular}
  \end{center}
\end{table}

Two conclusions follow. First, on a desktop-class CPU the algorithm does not reach real time
except in the most degraded settings, so it cannot meet the far tighter budget of an SBC.
Second, and more instructive, the accuracy is not monotone in the amount of computation
spent: $480$p at $30$\,fps overcounts ($59$) while $120$p undercounts badly ($30$), because
reducing resolution and frame rate degrades the tracker's data association before it
degrades the detection itself. Cheapening a tracking-based counter does not gracefully trade
accuracy for speed --- it breaks it. That asymmetry is the argument for moving the decision
out of the tracker.

\section{Geometric counting without tracking}
\label{sec:method}

\begin{figure*}[h!]
\centering
\includegraphics*[width=0.9\textwidth]{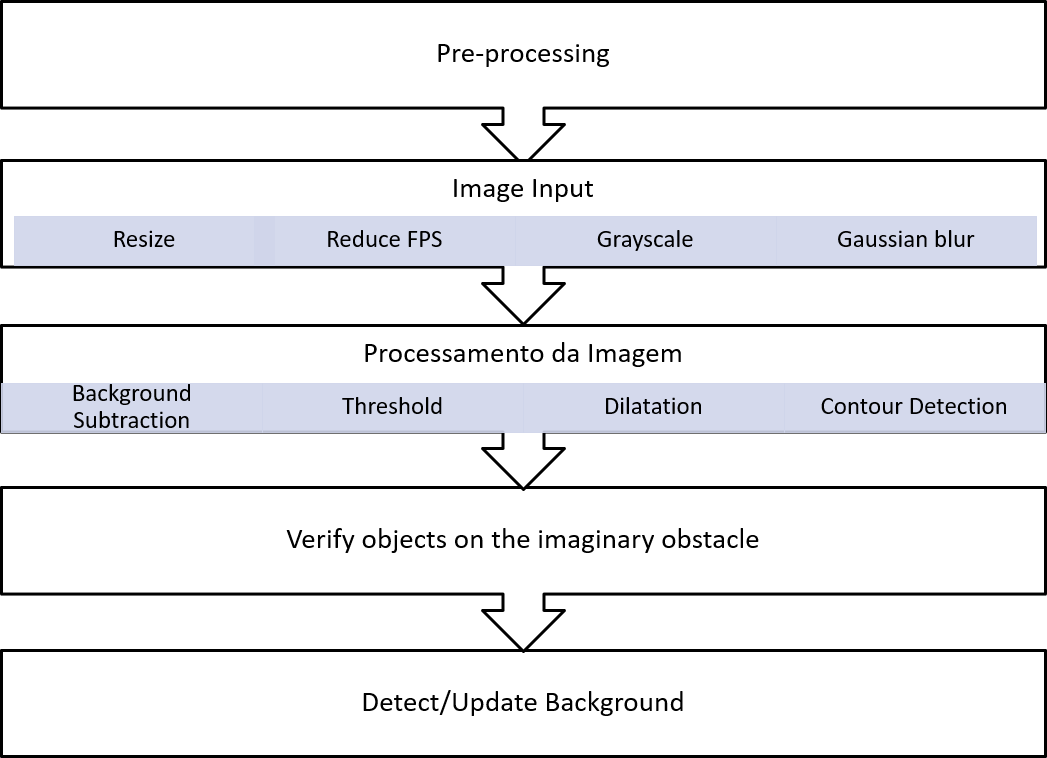}
\caption{Geometric detection method. The decision is attached to a fixed band in the image,
not to the detected objects, so no inter-frame state about objects is kept.}
\label{fig:met1}
\end{figure*}

Even when the search is limited to a band of interest, storing state from
previous frames and propagating it still costs more than the target platform can afford. The
reformulation is to change what the geometry is attached to: instead of following each
contour and asking where it will be, we attach the decision to the \emph{imaginary obstacle}
and ask whether anything is on it, regardless of what that something is. This is precisely
what an intrusive loop detector does, and it removes the tracker from the loop entirely. The
resulting pipeline is shown in Fig.~\ref{fig:met1}.

The front-end stages are unchanged, with aggressive resizing and frame-rate reduction to cut
processing time; what changes is the counting rule. Two rules are presented: the
\emph{constant average speed} strategy and the \emph{pre-calibration} strategy. The
fundamental difference with respect to tracking is that per-frame state collapses from a set
of object trajectories to a fixed-size occupancy vector with one entry per lane. The counter
never knows how many objects exist, where they came from or where they are going --- and for
the measurement of interest it does not need to.

\subsection{First rule: constant average speed}
\label{sec:alg1}
\begin{figure*}[h!]
\centering
\includegraphics*[width=0.9\textwidth]{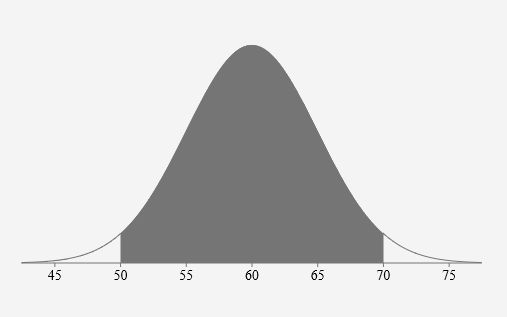}
\caption{Normal speed distribution assumed for the analysis, $\mu=60$\,km/h,
$\sigma=5$\,km/h. The shaded region is the $[50,70]$\,km/h interval, $95.45\%$ of the mass.}
\label{fig:normal}
\end{figure*}

The simplest verification rule assumes that all vehicles move at a similar average speed.
The assumption is not always true, and the consequences of its failure can be quantified.
Taking a speed distribution with mean $\mu = 60$\,km/h and standard deviation
$\sigma = 5$\,km/h, Fig.~\ref{fig:normal} shows the corresponding normal curve.

The shaded area is the probability that a passing vehicle travels between $50$ and
$70$\,km/h, namely $95.45\%$. Since the algorithm reasons about the centre of the blobs,
which move in units of pixels per frame, those speeds must be converted. In the area of
interest of the example video, $1$\,m corresponds to about $10$ pixels --- an equivalence
that depends on the resolution and on the region chosen --- and the frame rate is $30$, so
each second is $30$ frames. The conversion is

\begin{equation}
    V\!\left(\tfrac{\text{pixels}}{\text{frame}}\right)
    = \frac{10\,(\text{pixels}/\text{m})}{30\,\text{fps} \times 3.6}\;
      V\!\left(\tfrac{\text{km}}{\text{h}}\right).
\end{equation}

Speeds of $50$ and $70$\,km/h thus become $4.62$ and $6.48$ pixels/frame. Because pixel
displacements are discrete, at any given instant these must be treated as integers; rounding
gives $5$ and $6$ pixels/frame.

Let all vehicles be assumed to travel at $X = 6$ pixels per frame, the larger of the two. The
value of $X$ depends on resolution and frame rate, and rescales as

\begin{equation}
    X_{\text{new}} = X_{\text{old}}\;
    \frac{\text{Resolution}_{\text{new}}}{\text{Resolution}_{\text{old}}}\;
    \frac{\text{FPS}_{\text{old}}}{\text{FPS}_{\text{new}}},
    \label{eq:scaling}
\end{equation}

since the displacement per frame grows with the pixel--metre scale and shrinks with the
sampling rate.

\begin{figure*}[h!]
\centering
\includegraphics*[width=0.9\textwidth]{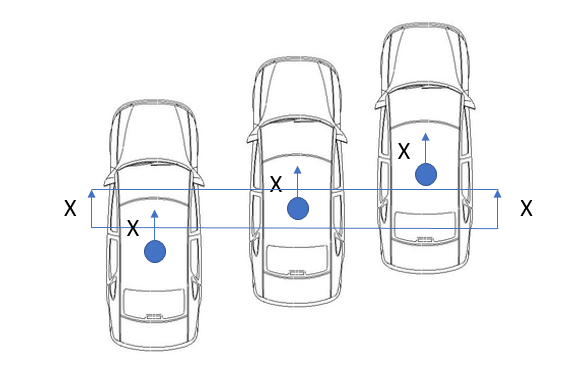}
\caption{Counting strategy based on the average speed. Band thickness is matched to the
expected inter-frame displacement.}
\label{fig:X}
\end{figure*}

The value of $X$ is the mean displacement per frame, and it is also chosen as the thickness
of the imaginary band drawn across the road. The choice is deliberate: if a centre inside
the band on one frame is outside it on the next, then the band thickness equals the
inter-frame displacement. Fig.~\ref{fig:X} shows the arrangement.

\begin{figure*}[h!]
\centering
\includegraphics*[width=0.9\textwidth]{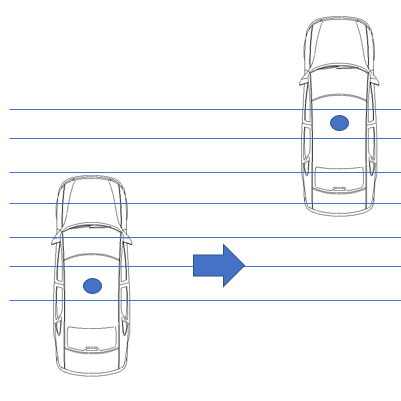}
\caption{Double-counting case: a vehicle slower than the design speed remains inside the
band for two consecutive frames.}
\label{fig:Y}
\end{figure*}

The rule is exact if every centre moves at exactly $X$. About half the vehicles, however,
travel at $Y = 5$ pixels/frame. With a band $6$ pixels long, a centre at speed $Y$ entering
the band at its first pixel is at pixel $6$ on the following frame, and is therefore counted
twice; Fig.~\ref{fig:Y} illustrates the case, each track representing one pixel.

This happens only when the centre enters the band at the first position. Entering positions
two to five (it cannot enter position six) it is counted once, so the rule is correct for
$80\%$ of the vehicles at speed $Y$. As those are about half of the fleet, the conditional
probability of a correct count is $90\%$; combined with the $95.45\%$ probability of the
speed lying in the design interval, the expected accuracy is approximately $85.9\%$.

The derivation is approximate and rests on stated hypotheses, but it makes the failure mode
explicit and, more usefully, it exposes the sampling condition that governs the whole
family: with band thickness $w$ and inter-frame displacement $v$, misses occur when $v > w$
and double counts when $v < w$. Only $v = w$ is exact, and since $v$ varies across the fleet
while $w$ is fixed, a single band can never be exact for all vehicles. This is the intrinsic
limit of the first rule, and the motivation for the second. It is also the reason for the
frame-rate collapse reported in Section~\ref{sec:obs-fps}: as the frame rate falls, $v$ grows
by Eq.~\ref{eq:scaling} until vehicles jump over the band entirely between samples.

\subsection{Second rule: pre-calibration by lane occupancy}
\label{sec:alg2}
\begin{figure*}[h!]
\centering
\includegraphics*[width=0.9\textwidth]{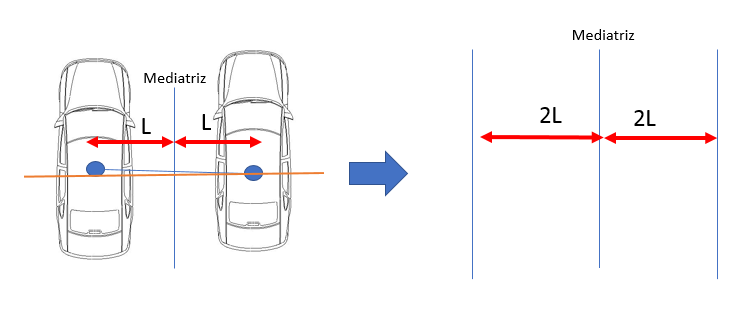}
\caption{Detection of lane lines from the statistics of the detected blob centres.}
\label{fig:lanes}
\end{figure*}

The limitation above is structural: the formulation itself embeds a known error. Together
with measurement error, that imprecision may be significant for the cases under study. It can
be removed by exploiting statistics that the front end already produces for free --- the mean
position of the centres of the vehicles that cross the sensor, their approximate area and the
mean width of the contours. From those it is possible to recover the approximate position of
the lanes, assuming that the mean centre of the vehicles travelling in a lane coincides with
the centre of the lane itself. The perpendicular bisector between two such mean centres is
the dividing line between the two lanes. Fig.~\ref{fig:lanes} illustrates the detection of
the lane lines.

\begin{figure*}[h!]
\centering
\includegraphics*[width=0.9\textwidth]{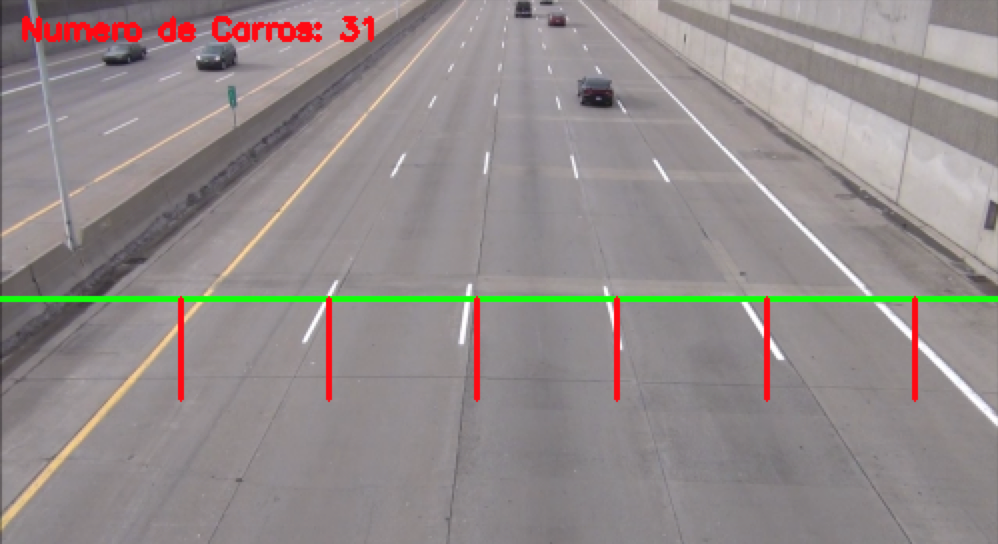}
\caption{Automated lane division derived from the pre-calibration statistics.}
\label{fig:lanes2}
\end{figure*}

This makes it possible to pre-calibrate a second rule based on the position of the lane
boundaries. The boundaries could be set by hand, but deriving them automatically is both more
accurate and more practical --- it is what allows the device to be pointed at a new road
without a survey. Dividing the road accordingly yields the regions of
Fig.~\ref{fig:lanes2}.

\begin{figure*}[h!]
\centering
\includegraphics*[width=0.9\textwidth]{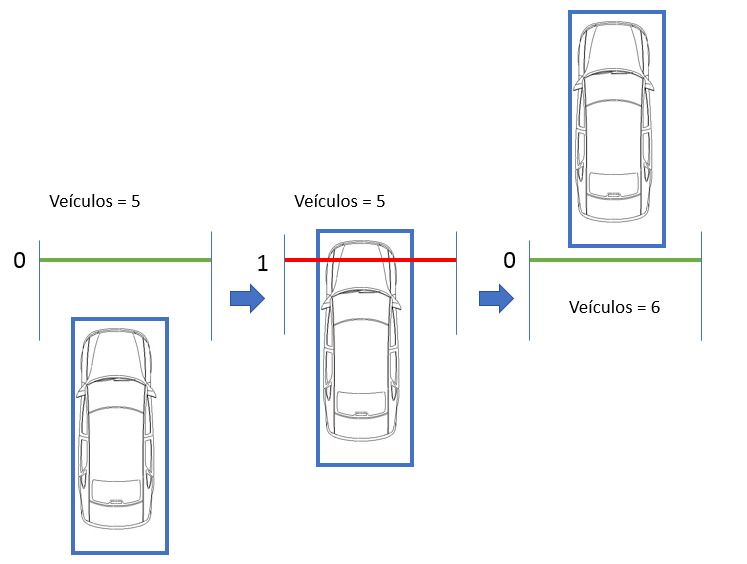}
\caption{Pre-calibration strategy: approach, occupancy and release of the lane counting
line.}
\label{fig:pre}
\end{figure*}

This second rule is called the \emph{pre-calibration} strategy. Counting proceeds in two
steps: if a lane presents a vehicle over the counting line, the entry corresponding to that
lane in the occupancy vector is set to one; when the lane no longer contains a vehicle in the
next captured frame, the entry returns to zero and one vehicle is added to the volumetric
count. In other words, counting is a falling-edge detector on a binary occupancy signal, one
signal per lane --- a software inductive loop~\cite{michalopoulos1991}, with the sampling
ambiguity of the first rule removed because the decision now depends on a state transition
rather than on a spatial coincidence. Fig.~\ref{fig:pre} shows a vehicle approaching, in
contact with the lane counting line, and leaving it.

The two rules have comparable computational cost. The advantage of the second is that it also
yields the average speed of the vehicles, since the number of frames a lane stays occupied is
the time the vehicle took to traverse its own length (Section~\ref{sec:speed}). It retains,
however, one failure mode inherited from the class-agnostic front end: two vehicles occupying
the same lane band back-to-back produce a single occupancy pulse and are counted once. No
appearance information is available to separate them, and this is the dominant source of
undercounting in dense traffic reported in Section~\ref{sec:observations}.

\section{Experiments}
\label{sec:results}

\subsection{Hardware characterisation}
\label{sec:hardware}
\begin{figure*}[h!]
\centering
\includegraphics*[width=0.9\textwidth]{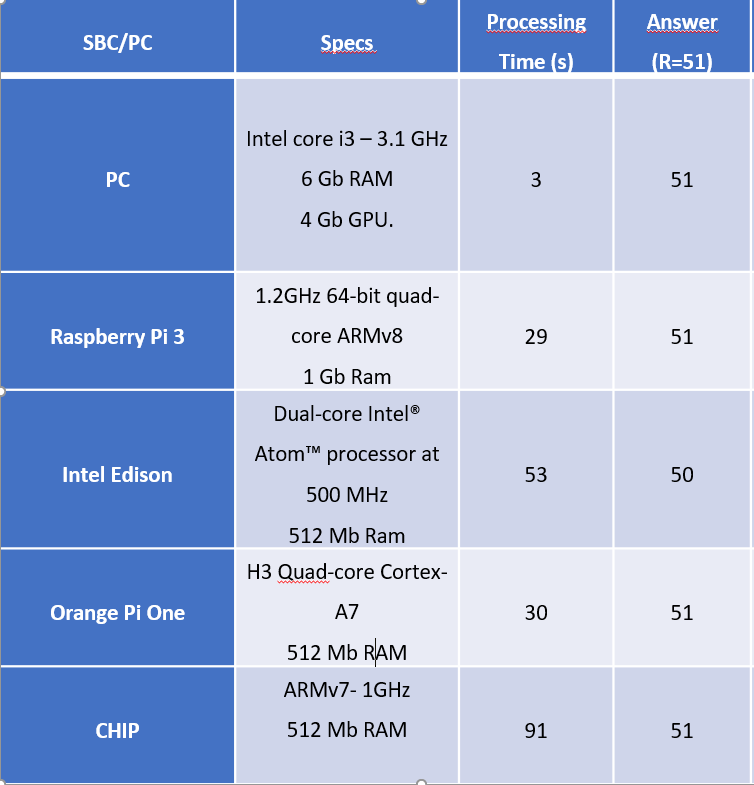}
\caption{Hardware characterisation across the candidate Single Board Computers and the
reference machine.}
\label{fig:hard}
\end{figure*}

The first test measures throughput on the candidate SBCs using the constant-average-speed
rule implemented in C++, on a 32\,s video at 240p and 10\,fps. The devices, their
specifications and the measured results are reported in Fig.~\ref{fig:hard}, together with
the reference desktop machine.

The counts were correct on every device except the C.H.I.P., where the error was small.
Both the Orange Pi and the Raspberry Pi run the algorithm faster than real time, which is
the requirement that the tracking baseline of Section~\ref{sec:baseline} failed to meet even
on a desktop CPU. The Raspberry Pi 3 was selected for the field deployment.

\subsection{Sweeps over resolution and frame rate}
\label{sec:sweeps}
\begin{figure*}[h!]
\centering
\includegraphics*[width=0.9\textwidth]{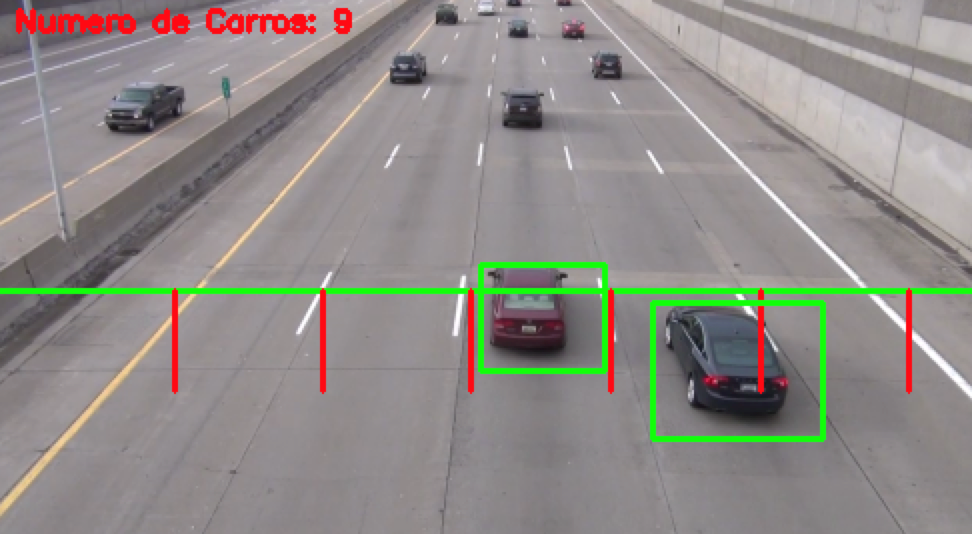}
\caption{Processed frame of the calibration video, showing the counting line, the recovered
lane division and the detected blobs.}
\label{fig:cali}
\end{figure*}

The first tests and calibrations used the video provided by Dahms~\cite{dahms2016}; a
processed frame is shown in Fig.~\ref{fig:cali}.

\begin{figure*}[h!]
\centering
\includegraphics*[width=0.9\textwidth]{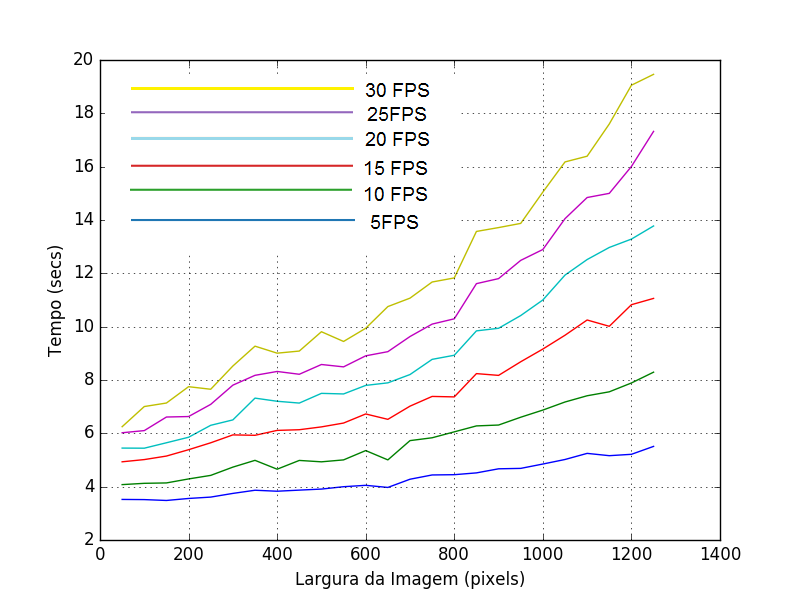}
\caption{Execution time of the geometric strategy as a function of image width, for several
frame rates.}
\label{fig:st}
\end{figure*}

\begin{figure*}[h!]
\centering
\includegraphics*[width=0.9\textwidth]{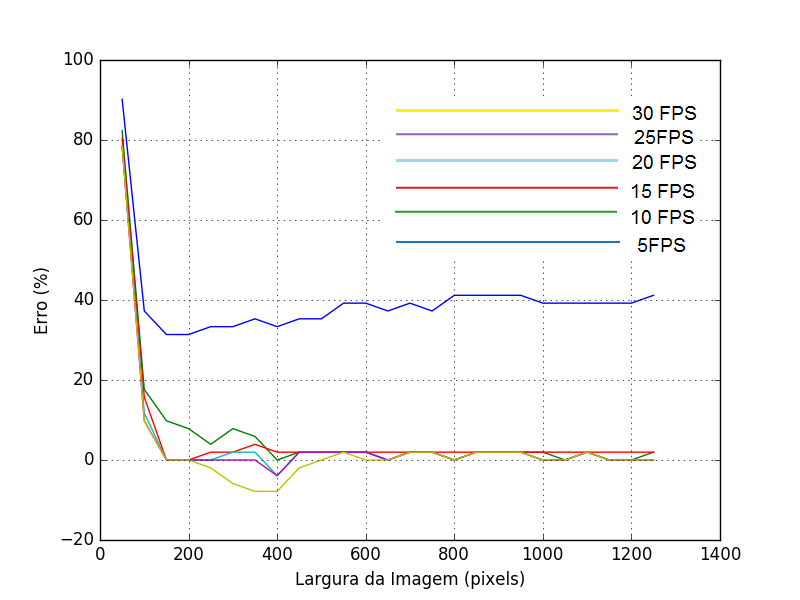}
\caption{Counting error of the geometric strategy as a function of image width, for several
frame rates. Note the knee near $200$ pixels and the error floor at $5$\,fps.}
\label{fig:se}
\end{figure*}

Both rules --- constant average speed (strategy 1) and pre-calibration (strategy 2) --- were
implemented in Python and evaluated over a grid of resolutions and frame rates. The image is
scaled equally in both directions and the width is used as the reference, swept from $50$ to
$1250$ pixels, the latter close to the $1280$-pixel width of HD. Execution time and counting
error after calibration are reported in Figs.~\ref{fig:st} and~\ref{fig:se}, with the error
defined as

\begin{equation}
    \text{Error} = 100\,\frac{N_{\text{real}} - N_{\text{algorithm}}}{N_{\text{real}}}\;[\%].
\end{equation}

Processing time grows with both frame rate and resolution, as expected. The error, in
contrast, does not behave smoothly, and the shape of the curves in Fig.~\ref{fig:se} is the
most useful result of the whole study. Error falls as resolution increases and rises as
frame rate decreases; above an image width of about $200$ pixels it settles very close to
zero and becomes insensitive to the frame rate used. Below that width it degrades quickly.
Separately, at a frame rate of five frames per second the error stabilises around $40\%$
instead of zero, at every resolution: at that sampling rate vehicles skip over the line of
interest between consecutive frames and are never observed on it. These two thresholds are
analysed in Section~\ref{sec:observations}.

\subsection{Multi-scene evaluation}
\label{sec:videos}
\begin{figure*}[h!]
\centering
\includegraphics*[width=0.9\textwidth]{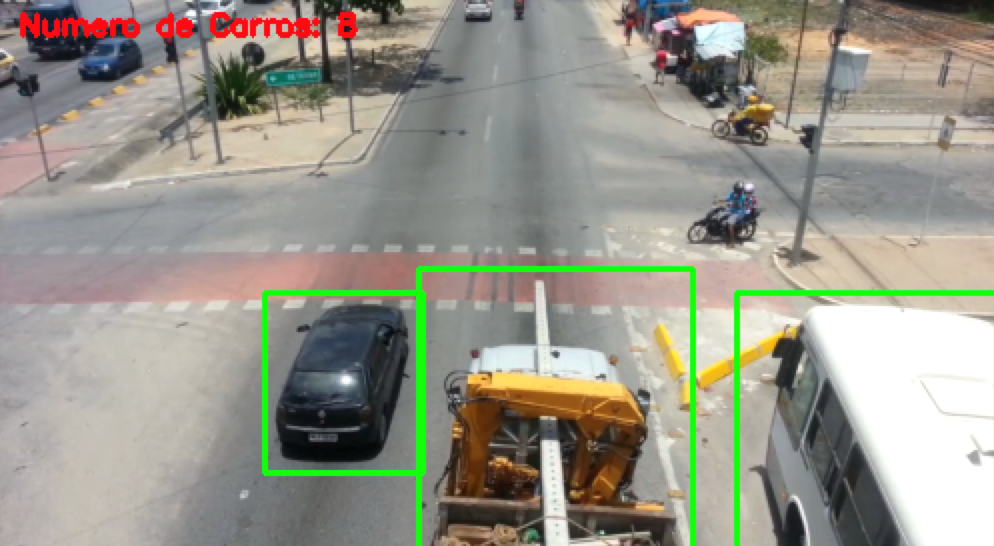}
\caption{Capture geometry of video 2: camera mounted on a pedestrian walkway above the
traffic.}
\label{fig:ufal}
\end{figure*}

The pre-calibrated algorithm was then evaluated on four videos from different locations. The
first is the clip released by Dahms, used in the performance tests above. The second was
captured on a road of interest with the camera placed on a pedestrian walkway above the
traffic, as illustrated in Fig.~\ref{fig:ufal}. The remaining two are publicly available
sample recordings from fixed surveillance cameras.

Table~\ref{tab:videos} reports the results of the pre-calibration algorithm, implemented in
Python, on each of the four videos.

\begin{table}[ht]
  \begin{center}
    \caption{Pre-calibration strategy (Python) on four videos from different scenes.
    Accuracy is $100\,(1-|N_{\text{real}}-N_{\text{alg}}|/N_{\text{real}})$.}
    \label{tab:videos}
    \begin{tabular}{ccccc}
      \toprule
      \textbf{Video} & \textbf{Duration} & \textbf{Real count} & \textbf{Algorithm count} & \textbf{Accuracy}\\
      \midrule
        1 & 32\,s & 51 & 51 & 100\%\\
        2 & 22\,s & 21 & 21 & 100\%\\
        3 & 35\,min & 2352 & 2014 & 85.6\%\\
        4 & 4\,min & 102 & 81 & 83.3\%\\
      \bottomrule
    \end{tabular}
  \end{center}
\end{table}

The split between the two short curated clips ($100\%$) and the two long uncontrolled
surveillance recordings ($83.3\%$--$85.6\%$) is discussed in
Section~\ref{sec:obs-generalisation}; both long-video errors are undercounts.

\subsection{Field deployment}
\label{sec:field}
The practical tests used the algorithm on a Raspberry Pi 3 on a real avenue, with the board
and camera positioned on a walkway above the traffic. Four configurations were compared
against the blob-tracking baseline: average vehicle speed and pre-calibration, each in
Python and in C++. Both language implementations were tested deliberately, to separate the
cost of the algorithm from the cost of the runtime. Communication with the device was
established over SSH from a mobile device, used to issue commands and collect the computed
data. Results are in Table~\ref{tab:field}. Each row corresponds to a different observation
window, hence the different ground-truth counts.

\begin{table}[ht]
  \begin{center}
    \caption{Field deployment on a Raspberry Pi 3 over a real avenue. Each row is an
    independent observation window with its own manual ground truth.}
    \label{tab:field}
    \begin{tabular}{cccc}
      \toprule
      \textbf{Algorithm} & \textbf{Real count} & \textbf{Algorithm count} & \textbf{Accuracy}\\
      \midrule
        Blob tracking (C++)     & 40 & 15 & 37.5\%\\
        Average speed (Python)  & 43 & 48 & 89\%\\
        Average speed (C++)     & 47 & 40 & 85\%\\
        Pre-calibration (Python)& 52 & 48 & 91\%\\
        Pre-calibration (C++)   & 45 & 35 & 77.8\%\\
      \bottomrule
    \end{tabular}
  \end{center}
\end{table}

The strategies developed here reach usable accuracy on the device. The blob-tracking
baseline~\cite{dahms2016} does not, precisely because the frame rate and resolution had to be
cut so aggressively for it to run on the prototype --- the failure predicted by
Table~\ref{tab:dahms}. Both geometric strategies performed better in Python than in C++,
which is not a statement about language speed but about tuning: the parameters and the
configuration were far easier to adjust in the Python implementation, and the accuracy
difference reflects how well each implementation was calibrated, not how fast it ran. The
runtime cost went the other way --- the Python implementations saturated the device at
$100\%$ CPU while the C++ ones used about $40\%$ --- which makes C++ preferable for long
unattended deployments, where sustained full load leads to overheating and thermal
throttling. Section~\ref{sec:obs-tuning} returns to this trade-off.

\subsection{Average speed as a by-product of occupancy}
\label{sec:speed}
\begin{figure*}[h!]
\centering
\includegraphics*[width=0.9\textwidth]{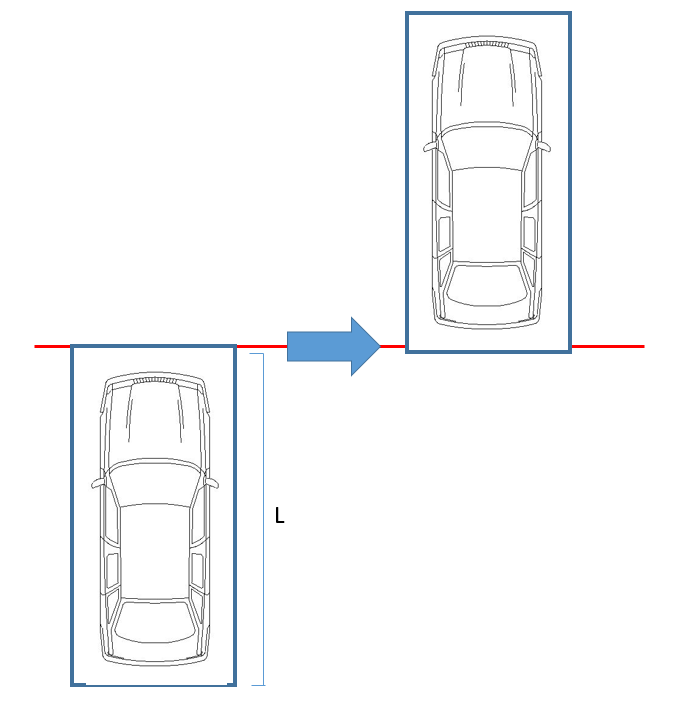}
\caption{Distance travelled by a vehicle while crossing the counting line. The occupancy
pulse width in frames is the traversal time of the vehicle's own length.}
\label{fig:speed}
\end{figure*}

The pre-calibration strategy also yields an estimate of the average speed of each vehicle at
the moment it is counted. Since the algorithm already registers the frame at which the
vehicle enters the imaginary counting line and the frame at which it leaves, the elapsed time
$\Delta T$ is available at no additional cost. Given the vehicle length $L$, estimated from
the bounding rectangle, that interval corresponds to the distance the vehicle travelled.
Fig.~\ref{fig:speed} illustrates the distance $L$ travelled while crossing the counting line.

The value of $L$, measured in pixels, is converted to metres using the lane width recovered
by the pre-calibration stage, whose real-world value is known:

\begin{equation}
L_{\text{metres}} = \frac{\text{Lane width}_{\text{metres}}}{\text{Lane width}_{\text{pixels}}}\; L_{\text{pixels}}.
\end{equation}

The average speed then follows from the ratio between $L_{\text{metres}}$ and $\Delta T$ in
seconds:

\begin{equation}
V_{\text{avg}}\,[\text{km/h}] = 3.6\;\frac{L_{\text{metres}}}{\Delta T\,[\text{s}]}.
\end{equation}

Applied to the first example video, this gives an average speed of approximately $70$\,km/h
for the passing vehicles. The estimate is coarse --- $\Delta T$ is quantised to the frame
period, so its relative error is $1/(\text{fps} \cdot \Delta T)$, which at the low frame rates
used here is substantial for a single vehicle --- but it is unbiased enough to be useful in
aggregate, and it costs one subtraction. The wider point is that the occupancy signal carries
more information than the count: pulse width gives speed, pulse area gives an apparent size
proxy, and the duty cycle of the lane signal is the lane occupancy rate, which is itself a
standard traffic-engineering quantity.

\section{Observations from the field}
\label{sec:observations}

This section collects what the experiments actually taught us. These observations were made
on 2016--2018 hardware, but none of them is a property of that hardware: they follow from the
sampling geometry and from the class-agnostic nature of the front end, and they still
constrain edge deployments built on the same principles.

\subsection{There is a resolution knee, and it is lower than expected}
\label{sec:obs-resolution}
Counting error is essentially zero for image widths above roughly $200$ pixels and degrades
sharply below it, independently of frame rate (Fig.~\ref{fig:se}). The knee is not about
image quality in any perceptual sense --- a $200$-pixel-wide frame is unrecognisably coarse
to a human --- but about whether a vehicle blob survives the threshold and dilation stages
with enough area to pass the shape filters. Once it does, extra resolution buys nothing and
costs quadratically.

The practical rule this yields is to operate just above the knee, and to find the knee
empirically per site rather than assuming it. Deployments that pick the resolution from the
camera's native mode, or from what the board can barely sustain, land on the wrong side of
this curve in both directions.

\subsection{Frame rate has a hard floor, and it is a sampling limit}
\label{sec:obs-fps}
At $5$\,fps the error does not degrade gracefully: it saturates near $40\%$ at every
resolution. The mechanism is aliasing. By Eq.~\ref{eq:scaling}, halving the frame rate
doubles the inter-frame displacement $v$; once $v$ exceeds the band thickness $w$, a vehicle
that was before the band on one frame is past it on the next and is never observed on it.
Resolution cannot compensate, which is exactly why the $40\%$ floor is flat.

This gives an explicit design constraint, and it is the single most transferable result of
the paper:

\begin{equation}
\text{fps} \;\ge\; \frac{V_{\max}\,[\text{m/s}] \cdot s\,[\text{px/m}]}{w\,[\text{px}]},
\label{eq:nyquist}
\end{equation}

where $s$ is the pixel--metre scale in the region of interest and $w$ the band thickness. Any
virtual-loop counter, then or now, that violates Eq.~\ref{eq:nyquist} loses vehicles no matter
how good its detector is. It is worth stating plainly because modern pipelines hit it too:
running an accurate detector at a low frame rate to save energy reproduces this failure
exactly, and the resulting undercount is often misattributed to the detector.

\subsection{Curated clips overestimate accuracy by a wide margin}
\label{sec:obs-generalisation}
The strategy scored $100\%$ on both short clips and $83.3\%$--$85.6\%$ on the two long,
uncontrolled surveillance recordings (Table~\ref{tab:videos}), with both long-video errors
being undercounts. The gap has three identifiable causes, all of them consequences of the
front end being class-agnostic:

\begin{itemize}
\item \textbf{Merged blobs.} In dense or slow traffic, two vehicles in the same lane touch in
the mask and produce a single occupancy pulse. With no appearance model there is nothing to
split them on. This is the dominant undercount.
\item \textbf{Shadows and illumination drift.} A cast shadow is a moving foreground region and
is indistinguishable from the object that casts it; over 35 minutes the sun moves, and a
running-average background model tracks slow drift but not sudden changes such as a passing
cloud. Shadow-aware background models~\cite{kaewtrakulpong2001} address part of this.
\item \textbf{Camera motion.} Poles and walkways vibrate. A few pixels of global motion
invalidate a per-pixel background model instantly and produce a burst of spurious foreground.
\end{itemize}

The lesson generalises beyond this method: any evaluation of a fixed-camera counter on short
curated clips is close to meaningless. Duration and uncontrolled conditions, not scene count,
are what expose the failure modes.

\subsection{Removing the tracker is what made the system deployable}
\label{sec:obs-tracker}
Under an equal compute budget on the Raspberry Pi, the blob-tracking baseline scored $37.5\%$
and the geometric rules $77.8\%$--$91\%$ (Table~\ref{tab:field}). The reason is not that
tracking is a bad idea --- it is that tracking degrades discontinuously under resource
pressure. A tracker starved of frames loses data association, and a broken association
produces both misses and duplicates, which is why Table~\ref{tab:dahms} shows the baseline
overcounting at one setting and undercounting at another. The geometric rule has no
association step to break; its error is bounded by the sampling condition and is therefore
predictable and, crucially, \emph{correctable by calibration}.

Stated generally: on constrained hardware, prefer estimators whose failure mode is a bias one
can measure over estimators whose failure mode is a collapse one cannot.

\subsection{Tuning effort dominated language choice}
\label{sec:obs-tuning}
The Python implementations were more accurate than the C++ ones ($91\%$ vs.\ $77.8\%$ for
pre-calibration), while consuming $100\%$ of the CPU against $40\%$. Neither number is about
the languages. The accuracy difference is that thresholds, kernel sizes and shape filters
were far quicker to iterate on in Python, so the Python configuration was simply better
calibrated; the CPU difference is real and matters for a different reason, namely heat. A
passively cooled board held at $100\%$ CPU on a pole in the sun throttles, and a throttled
board drops frames, which by Eq.~\ref{eq:nyquist} silently converts a thermal problem into a
counting error.

The deployment conclusion we drew --- prototype and calibrate in the high-level language,
then port the \emph{tuned} parameters to the compiled implementation for unattended
operation --- is unchanged today, and the failure it prevents (thermal throttling quietly
degrading accuracy) is still one of the least instrumented failure modes in edge vision.

\subsection{The cheap sensor was the useful sensor}
\label{sec:obs-cheap}
Both Orange Pi and Raspberry Pi ran faster than real time (Section~\ref{sec:hardware}), which
means the marginal cost of an additional measurement point was the price of a board and a
camera. For an origin--destination matrix, coverage matters more than per-site precision: ten
sites at $85\%$ accuracy constrain the matrix far better than one site at $99\%$. This
argument for many cheap class-agnostic sensors over few expensive ones is, if anything,
stronger now that the alternative involves an accelerator, a thermal solution and a model
lifecycle per site.

\section{Retrospective: where class-agnostic background subtraction still wins}
\label{sec:retrospect}

Nothing above argues against learned detectors. Where a trained model for the target class
exists and the power budget allows it, detection plus association
\cite{redmon2016yolo,zhang2022bytetrack} dominates this method on every metric. The argument
concerns the complement of that condition, which in applied work is larger than the
literature suggests.

\subsection{The conditions under which motion beats recognition}

\begin{enumerate}
\item \textbf{No model, and no data to train one.} The target class is outside the vocabulary
of any available detector --- livestock on a rural road, dropped cargo, agricultural or mining
machinery, informal transport modes, or the open-ended category ``anything that entered this
area''. Background subtraction defines foreground by change rather than by recognition, so it
has no vocabulary to be outside of. This is the situation in which we still reach for it
first.
\item \textbf{Cold start.} Even when a detector is the end goal, annotated crops are needed to
train it. A motion-based front end run over archived footage produces those crops
automatically: it proposes regions, a human labels a few thousand of them, and the resulting
detector eventually replaces the proposer. Background subtraction is the bootstrap, not the
destination.
\item \textbf{Power and thermal budgets.} The pipeline of Section~\ref{sec:pipeline} is a
handful of arithmetic operations per pixel with no accelerator, no model weights and a working
set of a few frames. That is a different order of magnitude from any neural detector, and it
is what allows continuous operation on solar power or on a passively cooled board.
\item \textbf{Privacy by construction.} The system's internal representation is a binary mask
and a list of bounding rectangles. No identity is computable from it, and frames need never be
stored or transmitted. Meeting the same guarantee with a detector-based pipeline requires
policy and auditing; here it is a property of the representation.
\item \textbf{Scene-specific adaptation.} A background model fitted online to one fixed camera
adapts continuously to that camera's viewpoint, lighting and weather. A detector trained
elsewhere carries a domain gap that no amount of inference-time compute closes. For permanent
installations this asymmetry is real and often underestimated.
\item \textbf{Class-agnostic tracking.} The association stages of SORT~\cite{bewley2016sort},
the IOU tracker~\cite{bochinski2017iou} and ByteTrack~\cite{zhang2022bytetrack} consume boxes
and are indifferent to where the boxes came from. Feeding them blobs from a background model
yields a training-free multi-object tracker. It is weaker than its detector-based
counterpart, and it exists on day zero of a project.
\end{enumerate}

\subsection{The hybrid pipeline we use today}

In current practice the method of this paper is not used alone. The arrangement that has
proved durable is:

\begin{enumerate}
\item \textbf{Motion proposals.} A modern background model --- MOG2~\cite{zivkovic2004},
ViBe~\cite{barnich2011vibe} or SuBSENSE~\cite{stcharles2015subsense} rather than the running
average used here --- produces class-agnostic candidate regions at full frame rate and
negligible cost.
\item \textbf{Geometric gating.} The virtual-loop logic of Section~\ref{sec:alg2} decides
which candidates matter, subject to the sampling constraint of Eq.~\ref{eq:nyquist}. Most
frames terminate here.
\item \textbf{Selective recognition.} Only the gated candidates --- a small fraction of the
pixels and a small fraction of the frames --- reach whatever recogniser is available: a small
classifier, an open-vocabulary detector~\cite{radford2021clip,liu2024groundingdino}, or a
promptable segmenter~\cite{kirillov2023sam,ravi2024sam2} for mask refinement or for
propagating a class-agnostic instance across frames. The expensive model is invoked on
demand, not per frame.
\item \textbf{Association.} Standard tracking-by-detection association over whichever boxes
the previous stages produced, evaluated with the usual
metrics~\cite{bernardin2008clearmot,luiten2021hota} and, for traffic specifically, against
benchmarks such as~\cite{naphade2021aicity}.
\item \textbf{Harvesting.} Gated candidates are archived as training crops. As the labelled
set grows, stage 3 gradually takes over from stage 1 --- for those classes that end up having
a model, and only those.
\end{enumerate}

Background subtraction has moved from being the detector to being the attention mechanism and
the data engine. That role is not a legacy niche: it is what keeps the compute budget bounded
and what makes the open-set case tractable at all.

\subsection{Limitations}

The results here should be read with their scope in mind. The evaluation covers four videos
and one field site, all daytime, all with an elevated near-nadir viewpoint, and ground truth
was obtained by manual counting; no per-object association metric was computed, since the
method produces no object identities. The analytical accuracy of Section~\ref{sec:alg1}
assumes a Gaussian, single-mode speed distribution and free-flow conditions, and does not
hold in congestion, where the merged-blob failure of Section~\ref{sec:obs-generalisation}
dominates. Night operation, heavy rain and strongly oblique viewpoints were not evaluated and
are expected to be substantially worse. The speed estimate of Section~\ref{sec:speed} is
quantised by the frame period and was not validated against an independent speed reference.

\section{Conclusion}

Computer vision applied to traffic data collection is well established --- licence-plate
extraction, vehicle detection, vehicle counting --- and volumetric counting on a road segment
in particular is a parameter of considerable importance for the theoretical models used in
urban planning.
 
This paper revisited a geometric, tracking-free method for volumetric vehicle counting on
Single Board Computers, in which detection is class-agnostic and the counting decision is
attached to a fixed band in the image rather than to the detected objects. Two rules were
presented: a constant-average-speed rule with an analytically derived accuracy of
$\approx\!86\%$, and a self-calibrating lane-occupancy rule that recovers the lane geometry
from blob statistics, counts occupancy edges and yields per-vehicle average speed as a
by-product. On four videos the second rule counted with $83.3\%$--$100\%$ accuracy, and in a
field deployment on a real avenue it reached $91\%$ against $37.5\%$ for a blob-tracking
baseline under the same compute budget, while running faster than real time on Raspberry Pi
class hardware.

The technical contribution that survives the intervening years is not the specific pipeline
but the constraints it exposed: a resolution knee below which foreground blobs stop surviving
morphological filtering, a frame-rate floor set by Eq.~\ref{eq:nyquist} below which objects
alias past the counting region regardless of detector quality, the discontinuous collapse of
association-based counting under resource pressure, and the thermal coupling between CPU
saturation and counting accuracy. These bind any edge vision system built on a fixed camera,
including those built on learned detectors.

What has changed is the role of the method rather than the method itself. Background
subtraction is no longer a competitive detector, and it was never meant to compete on
accuracy. What it offers is detection without a class model: it remains the tool of choice
when the object of interest has no trained detector, when annotated data does not yet exist,
when the power or privacy budget forbids a neural model, and when class-agnostic proposals are
needed to bootstrap the dataset that will eventually train one. In a research landscape
organised around closed-vocabulary benchmarks, that complement is easy to overlook and, in
deployed systems, remarkably common.

Future work follows the same line: replacing the running-average model with a modern
background model while leaving the decision logic unchanged, quantifying the merged-blob
failure explicitly against per-object ground truth, and characterising the hybrid gating
pipeline of Section~\ref{sec:retrospect} in terms of energy per counted object rather than
accuracy alone.

\bibliography{mybibfile}

\end{document}